\documentclass[PHME, 2022]{PHMSociety}

\usepackage{graphicx}
\usepackage{amsmath}
\graphicspath{{Figures/}} 

\begin{document}




\title{Detecting Manufacturing Defects in PCBs via Data-Centric Machine Learning on Solder Paste Inspection Features}

\author{%
	Jubilee Prasad Rao\authorNumber{1}, Roohollah Heidary\authorNumber{1}, and Jesse Williams\authorNumber{1} 
}

\address{
	\affiliation{{1}}{Global Technology Connection, Inc, Atlanta, GA, 30339, USA}{ 
		{{jrao,jwilliams,rheidary}@globaltechinc.com}
		} 
}

\maketitle
\pagestyle{fancy}
\thispagestyle{plain}


\begin{abstract}

Automated detection of defects in Printed Circuit Board (PCB) manufacturing using Solder Paste Inspection (SPI) and Automated Optical Inspection (AOI) machines can help improve operational efficiency and significantly reduce the need for manual intervention. In this paper, using SPI-extracted features of 6 million pins, we demonstrate a data-centric approach to train Machine Learning (ML) models to detect PCB defects at three stages of PCB manufacturing. The 6 million PCB pins correspond to 2 million components that belong to 15,387 PCBs. Using a base extreme gradient boosting (XGBoost) ML model, we iterate on the data pre-processing step to improve detection performance. Combining pin-level SPI features using component and PCB IDs, we developed training instances also at the component and PCB level. This allows the ML model to capture any inter-pin, inter-component, or spatial effects that may not be apparent at the pin level. Models are trained at the pin, component, and PCB levels, and the detection results from the different models are combined to identify defective components.

\end{abstract}

\section{Introduction}
\label{sec:intro} 

Printed Circuit Boards (PCBs) are an essential part of most electronic devices. Automated manufacturing of PCBs primarily includes solder paste printing, surface mount device placement, and a re-flow oven stage. It has been shown that about 50-70\% of PCB manufacturing defects are associated with solder paste printing process \cite{burr1997solder}. A Solder Paste Inspection (SPI) machine can be utilized after solder paste printing to capture information about the amount and location of solder paste deposited at each of the PCB pins.
It was found that a good correlation existed between SPI-extracted features and defects detected by an Automatic Optical Inspection (AOI) station located after the re-flow oven \cite{chintamaneni2015determination}. Several defect detection algorithms at SPI and AOI have been proposed in recent years that involve machine vision tools, image processing, and/or machine learning (ML) techniques to detect faults \cite{zakaria2020automated, khalilian2020pcb, rehman2019automated, wu2008real}. These works include models trained on a small number of PCB images ranging from 29 to 1500 PCBs, with and without defects. Since each PCB design may be different from another, these ML models may not be readily transferred over.

Rapid automated identification of faulty pins on a PCB at an early stage of PCB manufacturing, like SPI, could have significant cost savings to PCB manufacturers and consumers. A defective PCB could be reworked at one-tenth and one-fiftieth cost at the SPI stage compared to after the re-flow oven and after an in-circuit test respectively \cite{burr1997solder}. Hence, utilization of SPI stations along with algorithms that can ingest SPI features to accurately predict PCB defects can significantly improve manufacturing operations including reducing rework costs.

Machine Learning algorithms tend to perform well in learning complex interactions between features to perform a variety of classification and regression tasks. In the Industry 4.0 era, ML algorithms are becoming vital in performing functions such as fault detection, prediction, and prevention \cite{angelopoulos2019tackling}. In order to continue and improve operations, advanced ML techniques need to be widely adopted, as they are developed. In addition, a new paradigm in ML that shifts the focus from fine-tuning of ML models through extensive hyperparameter tuning to ensuring data quality is being considered \cite{garan2022data}. In this paper, we demonstrate the utilization of advanced ML algorithms and a data-centric approach to defect detection in PCB manufacturing.

This paper is structured as follows. In Sec. \ref{sec:datasets}, we describe the datasets and the objectives of the data challenge for which this work was conducted. We then outline our approach and provide details on data processing, feature and model selection, and evaluation steps in Sec. \ref{sec:approach}. Finally, an analysis of results from different trained models is given in Sec. \ref{sec:results} followed by conclusions in Sec. \ref{sec:conclusions}.

\section{Datasets and Problem Description}
\label{sec:datasets} 

\begin{figure}[t]
\centering
\includegraphics[scale=.8]{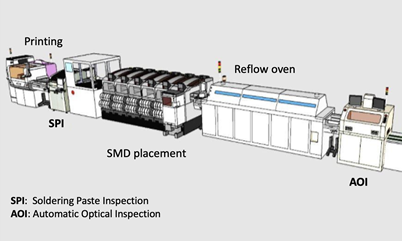}
\caption{PCB production line}  
\label{fig:production_line}
\end{figure}

Publicly available SPI feature dataset of 6 million PCB pins from a real-industrial production line are utilized in this effort. The production line is integrated with automated inspection machines as shown in Fig. \ref{fig:production_line}. The dataset also includes an AOI machine generated label and human operators given operator and repair labels. 

The provided datasets contained information pertaining to 15,387 identical PCBs with 8 PCBs per panel totalling 1,924 panels. A manufactured PCB had 128 components with different numbers of pins as listed in Table~\ref{table:pin_numbers}. The data provided features, extracted by the SPI machine, at the pin level for the nearly 6 million pins. The list of IDs and features generated by the SPI machine is shown in Table~\ref{table:SPI_features}. The SPI features include information about the amount and location of solder paste printed at for each pin on a PCB. The SPI machine also generated a ``Result" label that classified a pin as good or by the type of fault it predicted, such as insufficient solder, excessive solder, and others. Ninety-eight percent of the defects as detected by the SPI machine were labeled as ``W.Insufficient", and ``E.shape" and ``E.Position" were the labels for 0.8\% each of the SPI machine inspected pins. 


\begin{table}[t] \small  
	\begin{center}  
		
	\caption{Number of pins per component}
		\label{table:pin_numbers}
	
	\begin{tabular}{ c | c}
		\hline \hline
		\textbf{No. of components}	& \textbf{No. of pins}         \\ 
		\hline \hline
		108		&  2               \\ \hline
		1 	&         3    \\ \hline
		3 		&    5          \\ \hline
		7		&     6      \\ \hline
		8	&   8   \\ \hline
	    1	&    49   \\ \hline
	\end{tabular}
	\end{center}
\end{table}

\begin{table}[t] \small  
	\begin{center}  
		
	\caption{Features extracted by the SPI machine}
		\label{table:SPI_features}
	
	\begin{tabular}{ l | l| l}
		\hline \hline
		\textbf{Feature}	& \textbf{Description} & \textbf{Type}           \\ 
		\hline \hline
		
		PanelID		&  Panel ID         & ID                \\ \hline
		FigureID 	& Figure number in panel         & ID                \\ \hline
		Date 		& Date of mfg.          & Date              \\ \hline
		Time		& Time of mfg.          & Time              \\ \hline
		ComponentID	& Component name        & Categorical       \\ \hline
	    PinNumber	& Component pin number  & Categorical       \\ \hline
		PadID		& Panel pad number      & Categorical       \\ \hline
		PadType		& Type of pad (0 or 1)           & Categorical           \\ \hline
		Volume(\%)	& Volume of solder paste (SP)               & Numerical     \\ \hline
		Height(um)	& Height of SP              & Numerical     \\ \hline
		Area(\%)	& Area of SP                  & Numerical     \\ \hline
		OffsetX(\%)	& Offset X              & Numerical     \\ \hline
		OffsetY(\%)	& Offset Y              & Numerical      \\ \hline
		SizeX		& Size X                & Numerical     \\ \hline
		SizeY		& Size Y                & Numerical     \\ \hline
		Volume(um3)	& Volume in um3         & Numerical       \\ \hline
		Area(um2)	& Area in um2           & Numerical     \\ \hline
		Shape(um)	& Shape in um        	& Numerical     \\ \hline
		PosX(mm)	& Position X in mm      & Numerical       \\ \hline
		PosY(mm)	& Position Y in mm      & Numerical     \\ \hline
		Result		& SPI Result            & Categorical       \\ \hline
	\end{tabular}
	\end{center}
\end{table}

\begin{figure}[t]
\centering
\includegraphics[scale=.42]{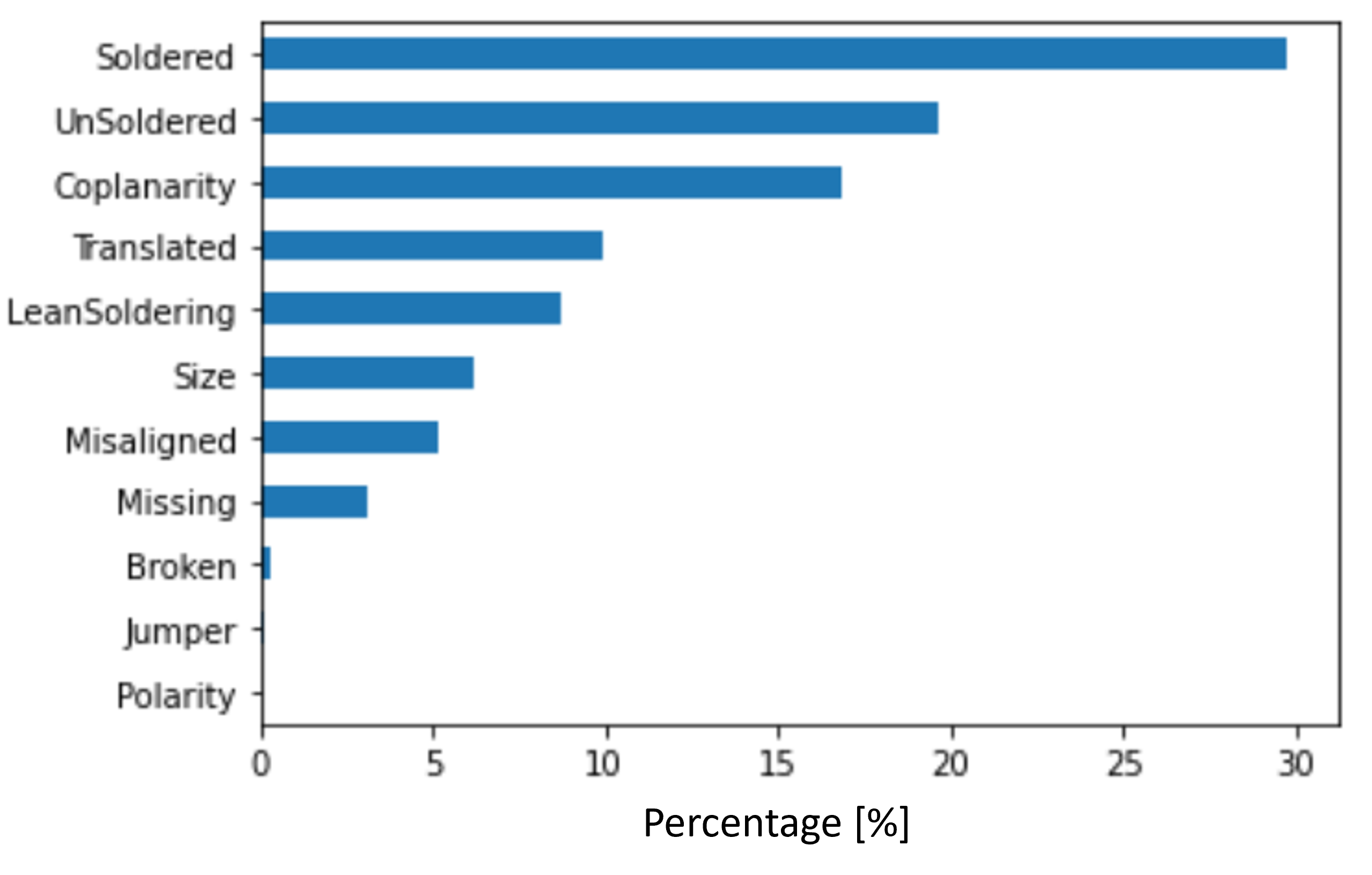}
\caption{Distribution of faults by type, labelled by AOI machines}
\label{fig:AOI_faults}
\end{figure}

Another file consisted only of pins found to be defective by the AOI machine. This dataset included labels given by the AOI machine as well as two human operators (``Operator Label", and ``Repair Label"). After the components are placed on the PCB and go through the re-flow oven, AOI machines utilize machine vision techniques to detect defects such as lean soldering, misalignment, and others. The distribution of the types of faults detected by the AOI machine are shown in Fig. \ref{fig:AOI_faults}. Each of the AOI-identified defective pins is then visually inspected by a human operator who gives an ``Operator Label" as ``Good" (not defective) or ``Bad" (defective). Further, another operator inspects the pins labeled as ``Bad" under a microscope, and adds a ``Repair Label" indicating whether it is ``False Scrap" (no defect) or ``Not Possible to Repair". In addition to the labels given by the AOI machine and the two human operators, this dataset also includes pin-identifying information (panel, figure, component, and pin IDs) needed to associate features of each pin in the AOI table to SPI features.


\begin{figure}[t]
\centering
\includegraphics[scale=.5]{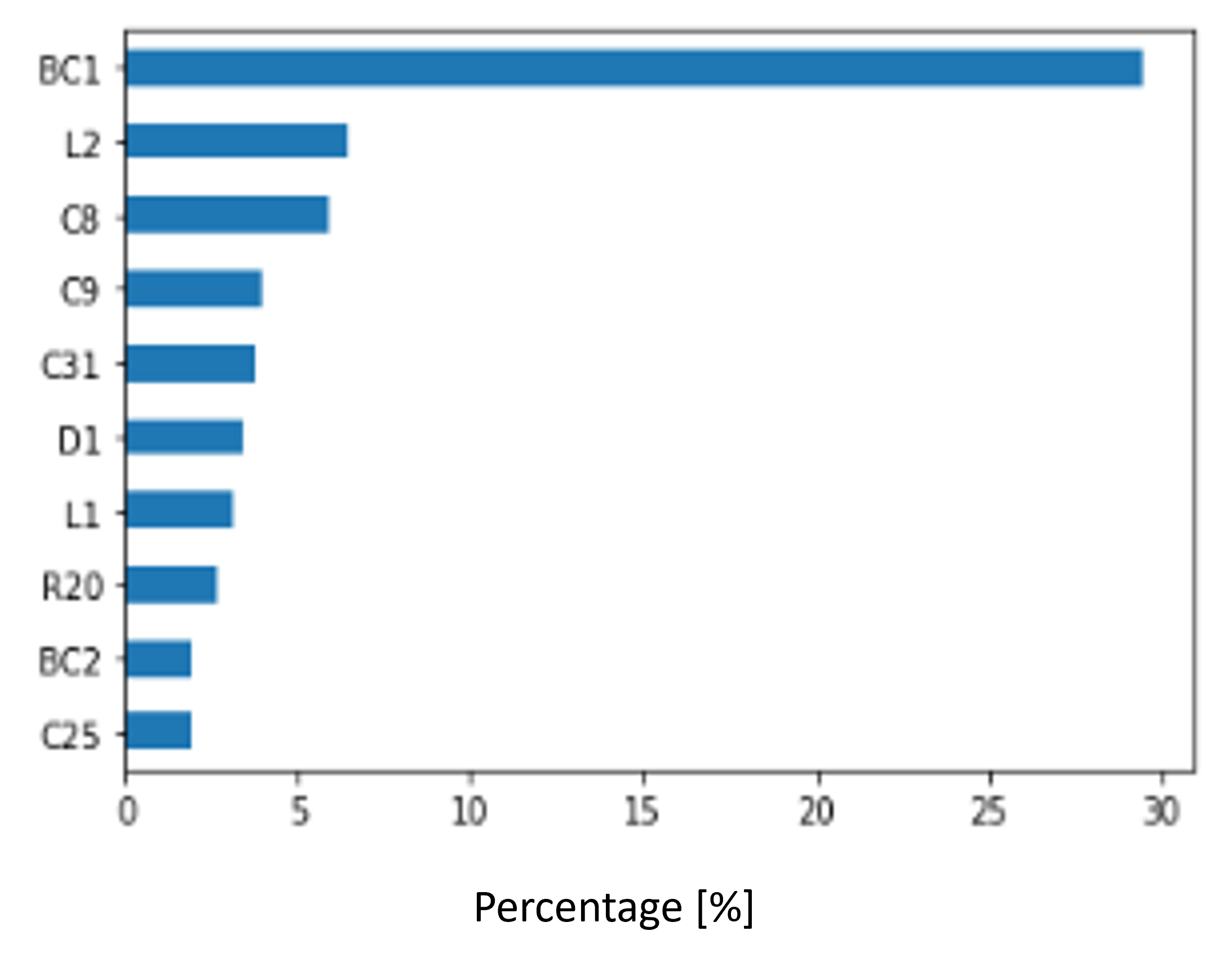}
\caption{Distribution of faults by component ID as labeled by AOI machines}
\label{fig:componentID_faults}
\end{figure}

\begin{figure}[t]
\centering
\includegraphics[scale=0.5]{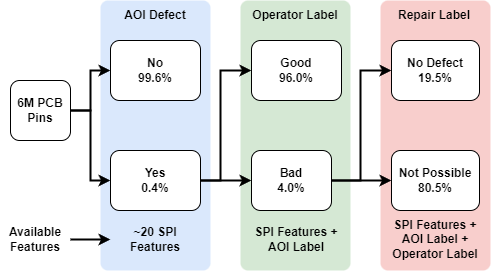}
\caption{Data availability for the three objectives}
\label{fig:challenges_flow}
\end{figure}

We have selected three labels as shown in Fig. \ref{fig:challenges_flow} to be predicted using ML techniques. The three objectives and the benefit each provides is listed below.

\begin{enumerate}
	\item Detect the presence of AOI defects using SPI features - To improve on current SPI fault detection algorithm
	\item Predict operator label using SPI features and AOI label - To automate operator label generation
	\item Predict repair label using SPI features and AOI and operator label - To automate repair label generation
\end{enumerate}

To evaluate the performance of the three algorithms, F1 score, given by Eq.~(\ref{eq:F1_score}), for the first prediction and macro average F1 score for the second and third predictions are utilized.

\textbf{REVIEW ABOVE SENTENCE OR SHOULD I JUST USE F1 SCORE FOR ALL PREDICTIONS OR I CAN CHANGE TO MCC}

\begin{equation}
F1 = \frac{2*TP}{2*TP+FP+FN}
\label{eq:F1_score}
\end{equation}

The ultimate goal of this classification system is to improve the operational efficiency of the PCB manufacturing process. A well-performing classification model can accurately identify the components to be reworked/eliminated after the SPI step, reducing the number of PCBs an operator has to inspect.

\section{Approach}
\label{sec:approach} 

\begin{figure}[t]
\centering
\includegraphics[scale=0.48]{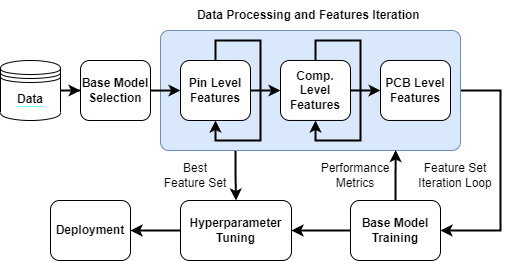}
\caption{The proposed data-centric ML approach}
\label{fig:data_centric_flow}
\end{figure}

We utilized a data-centric approach (Fig. \ref{fig:data_centric_flow}), where instead of using sophisticated ML algorithms, like for e.g., deep neural networks, and performing extensive hyperparameter tuning, we use relatively simple decision trees (XGBoost models) and concentrate on data pre-processing steps to achieve improved results. We developed models at three levels: at the 1. pin, 2. component, and 3. PCB levels. Pin-level models assume that all causes of a defective pin can be identified from its features alone. For component levels models, we expect that there may exist interactions between pins of a component that lead to a defective component. For the PCB level model, the interactions between components on a PCB are also considered. If the variation in solder quantity or impact of a pin defect is such that it almost never affects an adjacent pin or component, then the component and PCB level models may not provide additional value to the classification tasks. Before preparing training and test datasets for the three objectives, we performed cleaning of the data to rectify data formats and remove nan value rows, of which there were very few. Additional data processing and model training procedures are explained below.


\textbf{Features at the Pin, Component, and PCB Level:}
As mentioned before, the data was collected at the pin level with each pin's features stored as a row. Hence, to generate training datasets for the pin-level models, additional data processing was not needed. To generate component-level features to train the corresponding models, data was filtered according to the component ID. Then, if the component had two pins, the second-pin features were appended to the first-pin features in the same row. For components with additional pins, the process was repeated until all features for all the pins for that component were added to the row. This process was conducted for all components and resulted in 128 datasets, one for each component. Each of these datasets had 15,387 instances, equal to the number of PCB IDs available. The number of features in these datasets ranged from $\approx$ 40 in components with two pins to $\approx$ 1,000 features for the component with the most number of pins (49). Once the component-level features datasets are generated for all components, their corresponding models for the three objectives can be trained by using the pertinent target labels.

To generate PCB-level datasets, a similar process was followed, and all the component-level datasets were merged using the panel, figure, and PCB IDs. For the PCB level data, each row consisted of features for all its pins and for all components in a PCB. Here too, the total number of instances is equal to the number of PCB IDs but resulted in having just one dataset. This dataset had more about 5,000 features. Below, we describe the process of merging the features datasets with their corresponding labels for the three binary classification problems.

\textbf{Classification 1:}
For the first classification problem, we merged the SPI data with the AOI data using the panel, figure, component, and pin number IDs. Some of the instances (found to be defective by the AOI machine) in the AOI data were missing pin numbers, and since each component had more than one pin, this created ambiguity about which pin SPI features to associate such instances with. To avoid wrong associations, such AOI instances are not utilized while training the pin-level models by performing a left merge. This ambiguity does not exist at the component and the PCB level as pin number is not required to perform those merges. Since the objective is to detect the presence of a defect in a test component, all pins, components, or PCBs (based on the level) present in the AOI data were given a target of 1 versus 0 for all others.

\textbf{Classification 2:}
The objective of the second classification
is to determine if the fault detection, achieved by the AOI machine is accurate as determined visually by a human operator. Hence an inner merge is used, and only the pins, components, or PCBs depending of the data level that fail the AOI test are used as inputs for these models. The AOI label (lean soldering, translated, misaligned, or others) is added as an additional feature to the SPI features and the operator label (good - 0 or bad - 1) is used as the target label.

\textbf{Classification 3:}
For the third classification, the goal is to predict the repair label for those instances that have a ``bad" operator label given by a human visual inspector.
The operator labels are added to the SPI plus AOI label dataset as features, and the repair labels are replaced with 1 (not repairable) or 0 (no defect).

\textbf{Model Training and Evaluation}
Once the datasets at the three levels were generated, we used a five fold split and trained several XGBoost models. For the pin level models, one dataset captured all component defects. For the component level models, there was an option to train either one model for each component or one model for all the components combined. We trained models using both approaches. For the PCB level models, each set of features and target labels can only capture defects for one component at a time. This is because the targets are available at the component level. When component wise model training was needed, it was achieved using an iterative loop on the 128 components. 

When multiple models are used for classification, based on the assumption that higher-level models can capture inter-pin or inter-component interactions, a defective classification from even one model is reported as defective. Other approaches could be to use a majority voting approach or using the average of defect probabilities. In addition to F1 scores, Receiver Operator Curves (ROC) were also generated to visualize and compare model performance. 


\section{Results}
\label{sec:results} 

\begin{figure}[t]
\centering
\includegraphics[scale=0.6]{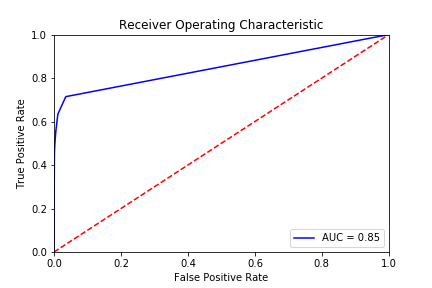}
\caption{ROC for pin level model to predict the presence of AOI pin defects}
\label{fig:ch1_pin}
\end{figure}

\begin{figure}[t]
\centering
\includegraphics[scale=0.6]{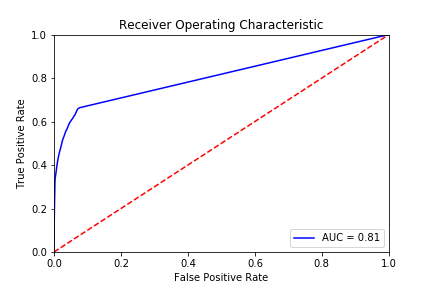}
\caption{ROC for the component level model to predict the presence of AOI component defects}
\label{fig:ch1_comp}
\end{figure}

\begin{figure}[t]
\centering
\includegraphics[scale=0.6]{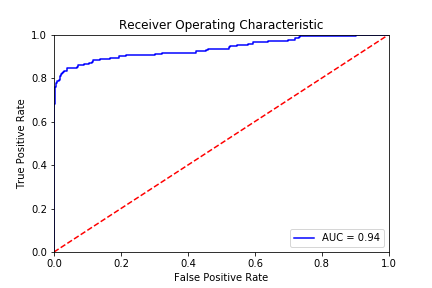}
\caption{ROC for pin level model to predict operator label}
\label{fig:ch2_pin}
\end{figure}

\begin{figure}[!h]
\centering
\includegraphics[scale=0.6]{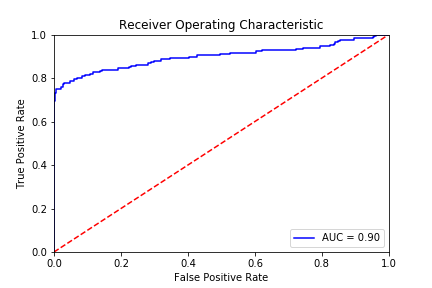}
\caption{ROC for the component level model to predict operator label}
\label{fig:ch2_comp}
\end{figure}

\begin{figure}[t]
\centering
\includegraphics[scale=0.6]{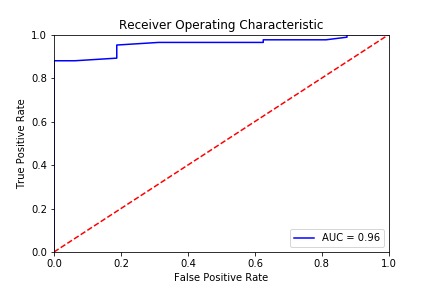}
\caption{ROC for pin level model to predict repair label}
\label{fig:ch3_pin}
\end{figure}

\begin{figure}[t]
\centering
\includegraphics[scale=0.6]{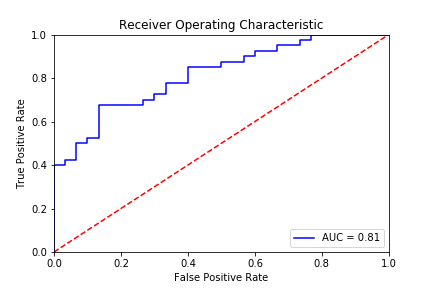}
\caption{ROC for the component level model to predict repair label}
\label{fig:ch3_comp}
\end{figure}

By following the approach described in Sec. \ref{sec:approach}, we trained nearly 400 classification models. The number of defects was very small compared to the total number of instances available. Only nearly 0.4\% of the about 6 million pin instances in SPI data were labeled as defective. Out of these only about 4.0\% of the instances were considered as defective by the visual inspector. Among these 80.5\% of the components were classified by the repair label as not possible to be repaired. These percentages are indicated in Fig. \ref{fig:challenges_flow}. Due to the very low number ($\approx 21,500$ for classification 1) of defective components, XGBoost models with a very small ($\approx 10$) number of features performed the best without over-fitting. The PCB-level models did not seem to improve classification performance. This may indicate that the deviation in the amount and location of printed solder paste from its designed values is not significant enough to cause inter-component defects. For classification problems 2 and 3, the number of components in each class was very small (including $\approx 150$) for PCB level ML models to learn from. It is observed that only 4.0\% of the number of SPI detected pin defects are classified as faulty by the AOI, and further nearly 20\% of the components labeled by AOI as defective were found to not be defective when inspected by an operator under the microscope (as given by the ``repair label"). This shows that there is significant room for improvement in these algorithms using machine learning approaches like the ones proposed here. The F1 scores for the different models evaluated on test data, for the three classification problems are tabulated in Table~\ref{table:results}. The depth of trees, which was found to be the most important factor, was only manually tuned. For most models trained, the depth of trees was set to 12 or less, and increasing it for the models did not result in and improvement of classification performance. This indicates that a defect at each pin or component is influenced by a very small number of extracted features given in Table~\ref{table:SPI_features}. ROC curves and Area Under the Curve (AUC) at the pin and component level models for the three challenges are shown in Figs. \ref{fig:ch1_pin} to \ref{fig:ch3_comp}. The lower performance of the component level model compared to the pin level model in classification problem 3 is also observed in Fig. \ref{fig:ch3_comp}. This may be due to the small number of training instances at the PCB level.

\begin{table}[t] \small  
	\begin{center}  
		
	\caption{F1 scores evaluated on test data}
		\label{table:results}
	
	\begin{tabular}{ c | c | c | c}
		\hline \hline
		\textbf{Ch.}	& \textbf{Model Level}   & \textbf{Components}  & \textbf{F1 Scores}     \\ 
		\hline \hline
		1	&  Pin level  	        & All       &  0.49     \\ \hline
		1	&  Component level      & All 	    &  0.55     \\ \hline
		1	&  PCB                  & Top 35    &  0.42     \\ \hline
		2	&  Pin level 	        & All    	&  0.80     \\ \hline
		2	&  Component level      & All       &  0.80     \\ \hline
		2 	&  PCB level            &  -   	    &    -      \\ \hline
		3	&  Pin level 	        & All       & 0.95      \\ \hline
		3	&  Component level      & All      	& 0.76      \\ \hline
		3	&  PCB level            &  -   	    &   -       \\ \hline
	\end{tabular}
	\end{center}
\end{table}

\begin{figure}[t]
\centering
\includegraphics[scale=0.6]{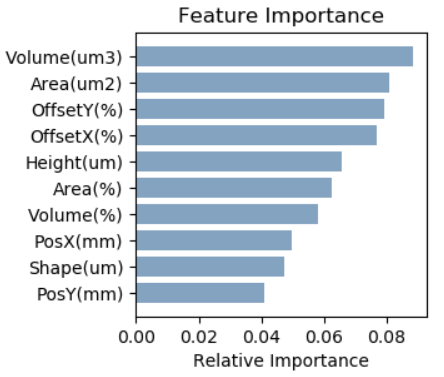}
\caption{Top ten important features to predict repair label as learned by a random forest ML model}
\label{fig:imip_features_rf}
\end{figure}

\section{Conclusions}
\label{sec:conclusions} 


A data-centric machine learning approach is utilized to detect PCB manufacturing faults at different stages of processing by considering features from each pin, component, and PCB at a time. This involved focusing more and improving the quality of the training data rather than training very complex models combined with extensive hyperparameter tuning. XGBoost algorithm is selected as the base and models are trained on the iteratively processed real production line data from 15,387 PCBs. The results show that improvements over currently utilized algorithms are possible and that significant reductions in manual inspections and rework costs are readily achievable using these models. F1 scores of 0.45, 0.8, and 0.73 are achieved to predict AOI defects, operator label, and repair label respectively. Further improvements through hyperparameter tuning of the machine learning models can be achieved. The data utilized to train the models is from only 9 days, and increasing the data size is expected to improve detection results. 


\section*{Acknowledgment}
The author would like to thank PHME society and Bitron Spa for publishing the utilized datasets. 

\section*{Nomenclature}

\begin{tabular}{ l  l }
	$PCB$			& printed circuit board         \\ 
	$SPI$			& solder paste inspection       \\ 
	$AOI$			& automated optical inspection  \\ 
	$XGBoost$		& extreme gradient boosting     \\ 
	$ML$			& machine learnng               \\  
	$TP$			& true positives                \\ 
	$TN$			& true negatives                \\ 
	$FP$			& false positives               \\ 
	$FN$  	   		& false negatives                \\
	$ROC$           & receiver operator characteristic \\
	$AUC$           & area under the curve              \\
 \end{tabular}

\bibliographystyle{unsrtnat}
\PHMbibliography{ijphm}

\end{document}